\documentclass[conference]{IEEEtran}

\usepackage[utf8]{inputenc}
\usepackage[T1]{fontenc}    
\usepackage{booktabs}   
\usepackage{url}

\usepackage[breaklinks]{hyperref}
\usepackage{amsfonts}  
\usepackage{hyperref} 
\usepackage{multirow}
\usepackage{amsmath}
\usepackage[numbers]{natbib}
\usepackage{graphicx}
\graphicspath{{./figs/}}
\usepackage{float}
\usepackage[table,xcdraw,dvipsnames]{xcolor}
\usepackage{subcaption}
\usepackage{booktabs}
\usepackage{comment}

\hypersetup{
    colorlinks=true,
    urlcolor=MidnightBlue,
    linkcolor=ForestGreen,
    citecolor=ForestGreen,
}
 
\usepackage[compact]{titlesec}
\titlespacing{\section}{1pt}{*0.5}{*0} 
\titlespacing{\subsection}{0pt}{*1}{*0}
\titlespacing{\subsubsection}{0pt}{*0}{*0}

\usepackage{array}
\newcolumntype{P}[1]{>{\centering\arraybackslash}p{#1}}
\newcolumntype{M}[1]{>{\centering\arraybackslash}m{#1}}

\usepackage[capitalize,
            nameinlink,
            noabbrev
            ]{cleveref}
  
\newcommand\blfootnote[1]{%
  \begingroup
  \renewcommand\thefootnote{}\footnote{#1}%
  \addtocounter{footnote}{-1}%
  \endgroup
}  

\font\myfont=cmr12 at 20pt
\title{\myfont Machine Learning Model Sizes and the Parameter Gap

\author{\IEEEauthorblockN{Pablo Villalobos\IEEEauthorrefmark{1}, Jaime Sevilla\IEEEauthorrefmark{1}\IEEEauthorrefmark{2}, Tamay Besiroglu\IEEEauthorrefmark{1}\IEEEauthorrefmark{3}, Lennart Heim\IEEEauthorrefmark{1}\IEEEauthorrefmark{4}, Anson Ho\IEEEauthorrefmark{1},  Marius Hobbhahn \IEEEauthorrefmark{1}\IEEEauthorrefmark{5}}
}}

\begin{document}
\bstctlcite{IEEEexample:BSTcontrol}
\maketitle

\begin{abstract}
\blfootnote{\IEEEauthorrefmark{1}Epoch, \IEEEauthorrefmark{2}University of Aberdeen, \IEEEauthorrefmark{3}MIT Computer Science \& Artificial Intelligence Laboratory, \IEEEauthorrefmark{4}Centre for the Governance of AI, \IEEEauthorrefmark{5}University of Tübingen}
We study trends in model size of notable machine learning systems over time using a curated dataset. From 1950 to 2018, model size in language models increased steadily by seven orders of magnitude. The trend then accelerated, with model size increasing by another five orders of magnitude in just 4 years from 2018 to 2022. Vision models grew at a more constant pace, totaling 7 orders of magnitude of growth between 1950 and 2022. 

We also identify that, since 2020, there have been many language models below 20B parameters, many models above 70B parameters, but a scarcity of models in the 20-70B parameter range. We refer to that scarcity as the \emph{parameter gap}.

We provide some stylized facts about the parameter gap and propose a few hypotheses to explain it. The explanations we favor are: (a) increasing model size beyond 20B parameters requires adopting different parallelism techniques, which makes mid-sized models less cost-effective, (b) GPT-3 was one order of magnitude larger than previous language models, and researchers afterwards primarily experimented with bigger models to outperform it. While these dynamics likely exist, and we believe they play some role in generating the gap, we don't have high confidence that there are no other, more important dynamics at play.
\end{abstract}


\section{Introduction}
\label{sec:introduction}

Over the past decades, we have witnessed a great increase in the size of machine learning models, measured in the number of free parameters that have to be fit to the data. Model size has become especially important as we have improved our understanding of scaling laws for language models, which govern how increases in model size and training data produce better performance \cite{kaplan2020scaling,Hoffmann2022}.

This implies that understanding trends in model size is relevant for understanding trends in performance. As an example, it is conceivable that scaling models beyond trillions of parameters proves to be a considerable challenge (for example, due to the lack of sufficient onboard memory in which to store the model weights). This would cause slower scaling of ML models going forward.

In this article, we study trends in the model sizes of milestone ML models from the 1950s to 2022. In section 2, we analyze the trends we have identified in the data. We find a significant speedup in model size growth starting from 2018, plus the apparition of a new trend of very big models starting from 2020. Section 3 provides a detailed discussion of the \emph{parameter gap} -- the striking scarcity of models of size between 20B and 70B parameters across all publication dates. We establish that the gap is unlikely to be the result of chance, and discuss five hypotheses to explain it. While we have been able to falsify some of the hypotheses, we do not have a definitive explanation.

\subsection{Previous work}
\label{sec:related-work}

To our knowledge, there are no previous studies on trends in model size of published machine learning systems across architectures. There have been previous efforts on gathering model size data \cite{talat-etal-2022-reap}. Similar analyses have been performed on compute utilization \cite{amodei2018compute}, as well as performance improvement over time \cite{papersWithCode}. In addition, the scaling literature contains many detailed reports on the properties of concrete architectures at different model sizes \cite{kaplan2020scaling,Hoffmann2022}. However, none of these address the study of trends in published model size across different architectures.

\begin{figure}[ht]
  \small
  \centering
  \includegraphics[width=0.45\textwidth]{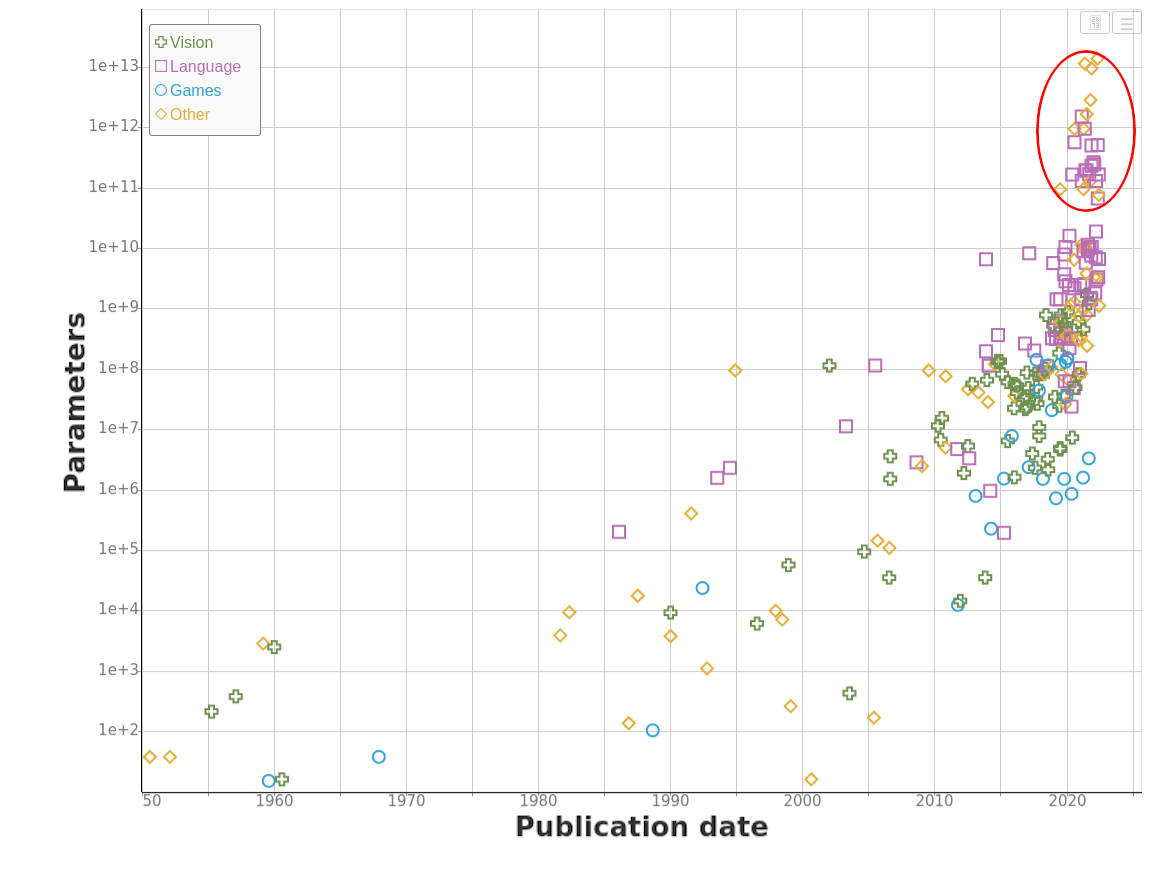}
  \caption{\small Full dataset, separated by category. The cluster of large models is indicated by a red circle.}
  \label{fig:flagship}
\end{figure}

\subsection{Our dataset}
\label{sec:dataset}

\begin{figure*}[ht!]
  \small
  \centering
  \includegraphics[width=0.45\textwidth]{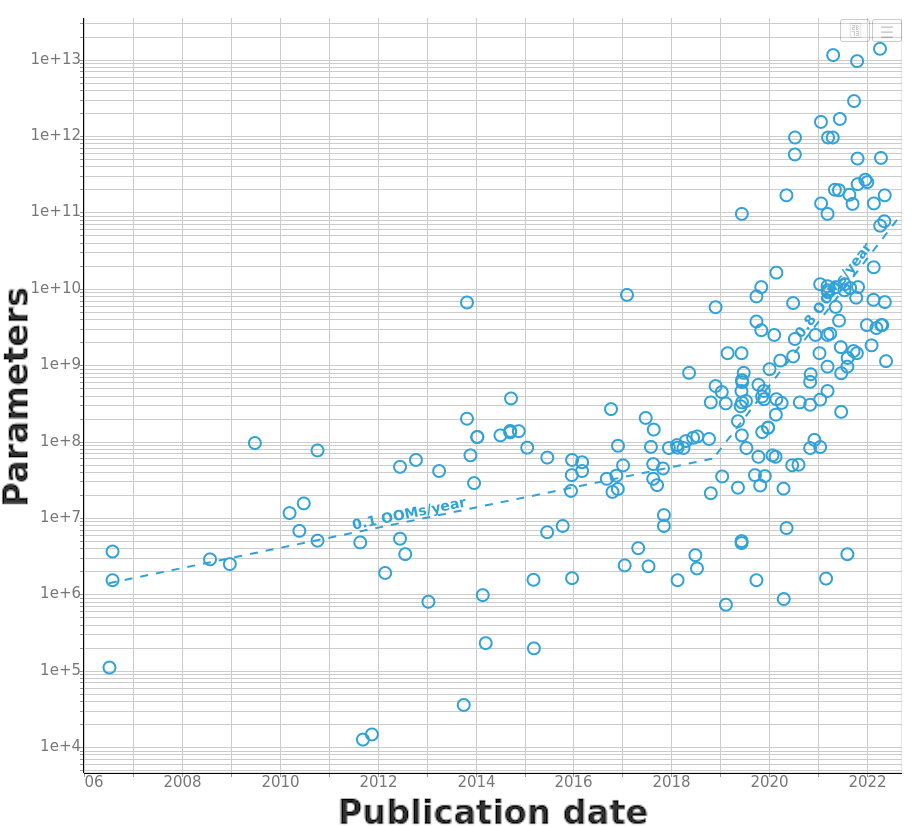}
  \includegraphics[width=0.45\textwidth]{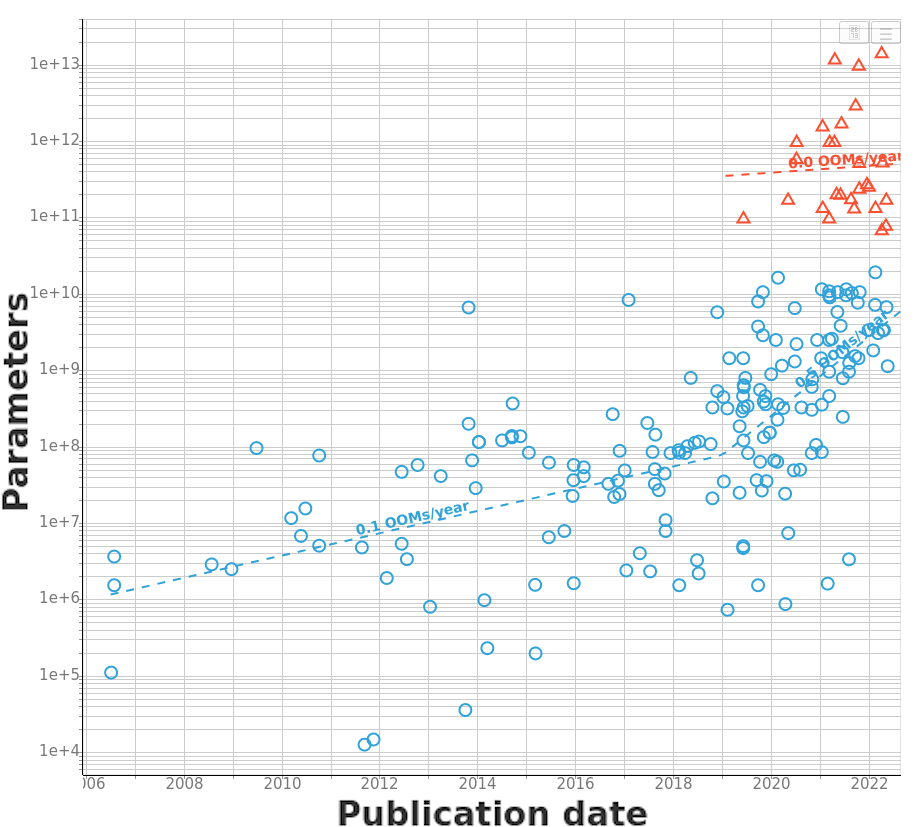}
  \caption{\small Log-linear regressions. Left: transition period between the pre-2018 and post-2018 trends. Right: Post-2018 trends, split by size}
  \label{fig:fig2}
\end{figure*}

In previous work, we have compiled a dataset of the parameter counts of 238 notable ML models \cite{sevilla2021parameters}, chosen based on a series of inclusion criteria. Specifically, we include models which satisfy at least one of the following:
\begin{itemize}
    \item Have more than 1,000 citations
    \item Have widely accepted historical importance
    \item Increased the state-of-the-art performance in a relevant benchmark or task
    \item Were deployed in an important context, such as a widely-used search engine or translation software.
\end{itemize}

Even if the total parameter count of a model is not specified in the accompanying paper, it is usually straightforward to determine this based on the model architecture. This is in contrast to calculating other metrics like training compute, where there is more uncertainty in the estimates.

Each model in our dataset is categorized into one of four domains according to the task it is designed to solve. The domains are \textit{Vision}, \textit{Language}, \textit{Games} and \textit{Other}\footnote{The \textit{Other} category compiles other domains like drawing, speech recognition or multimodal models. A domain is aggregated into \textit{Other} when we have less than 15 systems of that domain in the dataset.}. The full dataset can be seen in Figure~\ref{fig:flagship}. The data covers the period from the 1950s to 2022, but 80\% of the points are concentrated in the period after 2010, so it can only provide reliable information on the last decade of ML history.

\begin{table*}[ht!]
  \renewcommand{\arraystretch}{1.5}
  \centering
\begin{tabular}{@{}cccccc@{}}
\toprule
\textbf{Period}  & \textbf{Data}   & \textbf{Scale (start to end)} & \textbf{Slope}   & \textbf{Doubling time} & \textbf{R$^2$}  \\ \midrule
\rowcolor[HTML]{FCF7EB} 
\begin{tabular}[c]{@{}c@{}}1952 to 2018\end{tabular}     & \begin{tabular}[c]{@{}c@{}}$n=109$\end{tabular} & 1e+01 to 3e+7 params & \begin{tabular}[c]{@{}c@{}}0.1 OOMs/year \\ {[}0.1; 0.1; 0.1{]}\end{tabular} & \begin{tabular}[c]{@{}c@{}}39.1 months \\ {[}36.4; 39.7; 40.7{]}\end{tabular} & \begin{tabular}[c]{@{}c@{}}0.62\end{tabular} \\
\rowcolor[HTML]{EAF6FB}
\begin{tabular}[c]{@{}c@{}}2018 to 2022\\ (single trend)\end{tabular}         & \begin{tabular}[c]{@{}c@{}}$n=129$\end{tabular}    & 3e+7 to 2e+12 params           & \begin{tabular}[c]{@{}c@{}}0.9 OOMs/year \\ {[}0.9; 0.9; 1.0{]}\end{tabular} & \begin{tabular}[c]{@{}c@{}}4.2 months \\ {[}3.5; 4.0; 4.3{]}\end{tabular}  & \begin{tabular}[c]{@{}c@{}}0.31\end{tabular}    \\
\rowcolor[HTML]{FFEDEA} 
\begin{tabular}[c]{@{}c@{}}2018 to 2022\\ (above gap)\end{tabular} & \begin{tabular}[c]{@{}c@{}}$n=27$\end{tabular}  & 7e+10 to 2e+12 params           & \begin{tabular}[c]{@{}c@{}}0 OOMs/year \\ {[}-0.4; -0.1; 0.2{]}\end{tabular} & \begin{tabular}[c]{@{}c@{}}209 months \\ {[}-52.5; -14.2; 52.0{]}\end{tabular}  & \begin{tabular}[c]{@{}c@{}}0.00\end{tabular}  \\
\rowcolor[HTML]{FFEDEA} 
\begin{tabular}[c]{@{}c@{}}2018 to 2022\\ (below gap)\end{tabular} & \begin{tabular}[c]{@{}c@{}}$n=102$\end{tabular}  & 3e+7 to 2e+10 params           & \begin{tabular}[c]{@{}c@{}}0.5 OOMs/year \\ {[}0.4; 0.5; 0.5{]}\end{tabular} & \begin{tabular}[c]{@{}c@{}}8 months \\ {[}7.0; 8.0; 9.8{]}\end{tabular}  & \begin{tabular}[c]{@{}c@{}}0.25\end{tabular}  \\ \bottomrule
\end{tabular}
  \vspace{.5em}
  \caption{Summary of our main results. Around 2018 there was a general increase in growth. This can be split into the previous trend increasing its growth rate, and a separate cluster of very large models appearing on top.}
  \label{tab:table1}
\end{table*}

\section{Trends in model size}

The most salient feature is the recent increase in the rate of growth of model size, starting from mid-2018\footnote{There is one Language system with around 1e+10 parameters from 2014, Word2Vec(large), which we treat as an outlier since it's a word embedding and thus has very different characteristics than most other systems}. The growth rate before 2018 was 0.1 OOMs/year (order of magnitudes per year), whereas after 2018 the growth rate is 0.9 OOMs/year if we assume a single exponential trend (see Figure~\ref{fig:fig2}).

However, we think this might be better explained as two different trends. It is visually apparent in Figure \ref{fig:flagship} that in the 2018-2022 period the models can be divided into two clusters: those above 70B parameters and those below 20B parameters. The separation between these clusters is the “parameter gap” that we mentioned earlier. However, doing this separation runs the risk of overfitting, given the small sample size and that we are splitting the data across time and model size into three different groups.

We can think of the models below the gap as mostly continuing the previous trend, whereas the models above the gap constitute a separate new trend of big models. With this distinction, the growth rate for models below the gap is around 0.5 OOMs/year, while for those above the gap there’s no identifiable trend\footnote{Fitting a trendline produces an R$^2$ of 0.00.}. A summary of all the regressions can be found in Table~\ref{tab:table1}.

\begin{figure}[ht]
  \small
  \centering
  \includegraphics[width=0.45\textwidth]{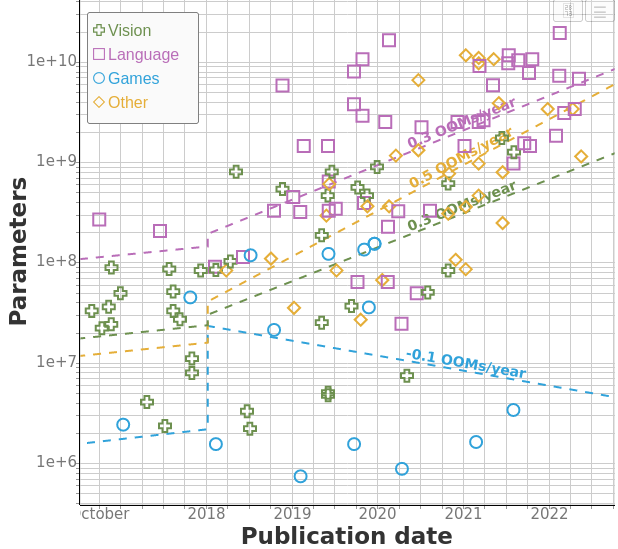}
  \caption{\small Below-gap trends by category.}
  \label{fig:category}
\end{figure}

If we look at the trends by category (see Figure~\ref{fig:category}), all of them were growing at 0.1 OOMs/year before 2018, but there are some differences after that. The \textit{Games} category seems to have been less prominent in recent years, slowing down its growth rate. Meanwhile, the \textit{Language} and \textit{Other} categories are leading the recent growth spurt, even after we remove the outliers above the parameter gap. The \textit{Vision} category sits in the middle of these two extremes, growing at a moderate pace.

Above the gap, we only see \textit{Language} and \textit{Other} systems. The \textit{Language} systems appear to be decreasing in size over time, but this is not really a significant trend and a few early big Mixture-of-Experts (MoE) systems might be causing this apparent decline\footnote{Since MoE models can be made much bigger than dense models. Concretely, Switch \cite{fedus_switch_2021} and GShard \cite{lepikhin_gshard_2020} are examples of early big MoE models that might distort the trend. Indeed, if we remove them the trendline looks completely flat for \textit{Language} systems.}.

It is perhaps surprising that this acceleration took place around 2018, since previous analyses on training compute trends found a similar increase in 2015 \cite{epoch2022computetrends}. This discrepancy might reflect the recent increased interest in natural language processing relative to other domains and the apparition of large language models, which have a relatively high size-to-compute ratio compared to other systems.

\section{The parameter gap} \label{sec:trends}

\begin{figure}[ht]
  \centering
  \includegraphics[width=0.45\textwidth]{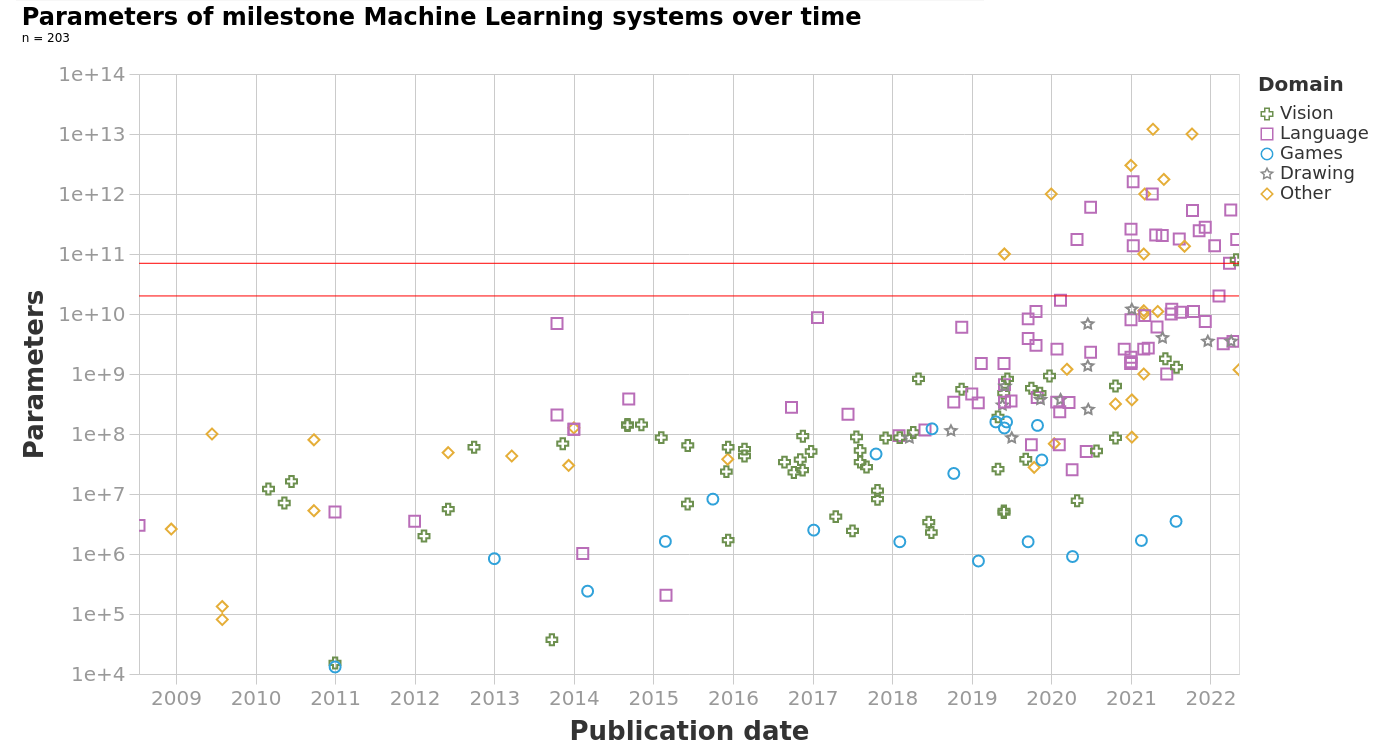}
    \vspace{0em} 
  \caption{\small The parameter gap, from 2+10 to 7e+10 parameters}
  \label{fig:gap}
\end{figure}

The parameter gap (Figure \ref{fig:gap}) was first observed by Jones 2021 \cite{jones2022}, who attributed it to differences in the computing infrastructure used to train the models. Here we will build on his observations to clarify the significance and nature of the gap. In section 3.1 we will establish that a gap of this size is unlikely to have been produced by chance. In section 3.2 we add some further context that we have found valuable to clarify possible explanations, and in section 3.3 we present our best hypotheses to explain the gap.

\vspace{0em}

\subsection{Statistical analysis}

Before attempting to explain this gap, we performed a simple analysis to try to establish the likelihood that this gap is due to random chance. We model the model size as an exponential trend with lognormal noise.

\begin{equation*}
    \log(s) = \beta t + \epsilon
\end{equation*}

where $t$ is the year, $s$ is the model size and $\epsilon \sim N(0, \sigma^2)$.

\begin{figure}[h]
  \centering
  \includegraphics[width=0.45\textwidth]{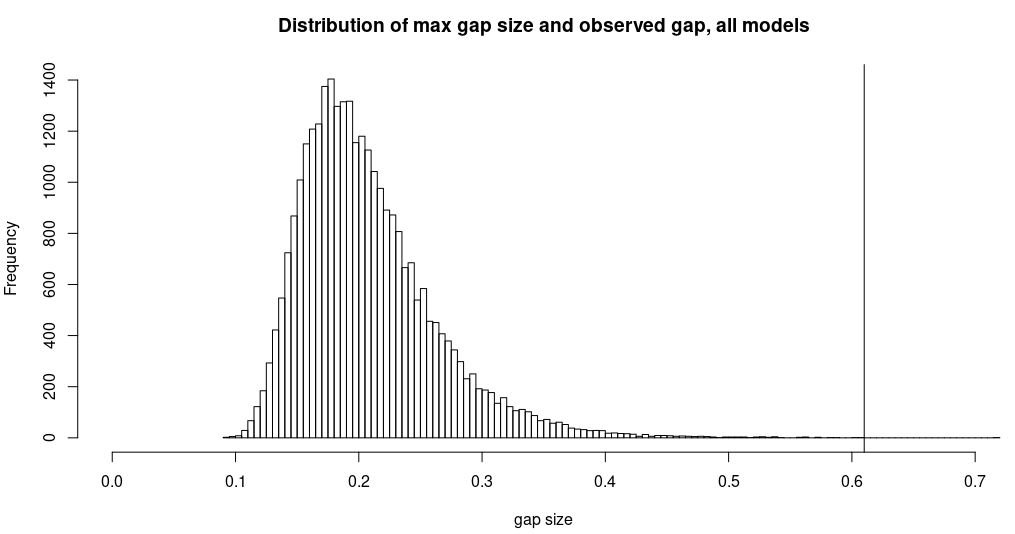}
    \vspace{0em} 
  \caption{\small Distribution of the widest parameter gap (in OOMs) for 10,000 simulations and observed gap size.}
  \label{fig:dmax}
\end{figure}

After fitting the trend slope ($\beta$) and the noise variance($\sigma^2$) to the observed data, we ran a Monte Carlo simulation of the model. In each simulation we sample 238 points from these distributions on the same dates as our observed data and calculate the maximum observed gap between two groups of at least 10 points. The resulting gap size distribution after 10,000 simulations, together with the observed gap, can be seen in Figure~\ref{fig:dmax}.

\vspace{0em}

The observed gap is extremely unlikely according to our simple model – the probability of observing a gap as large as the one present is on the order of $10^{-5}$. This strongly suggests that the gap is not simply due to random chance.

    \begin{figure*}[ht!]
      \small
      \centering
      \includegraphics[width=0.42\textwidth]{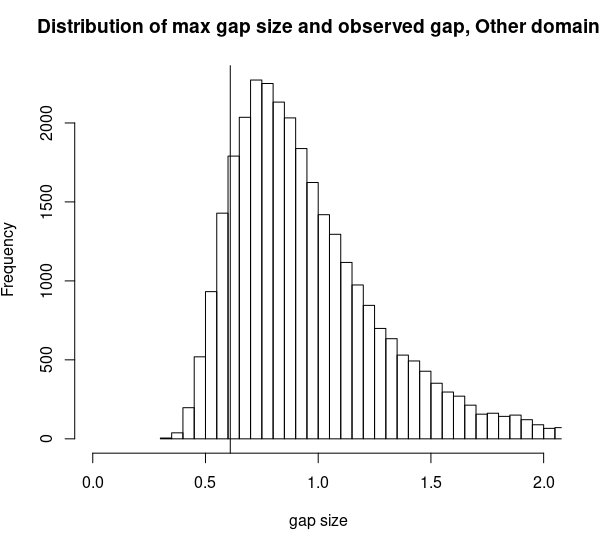}
      \includegraphics[width=0.48\textwidth]{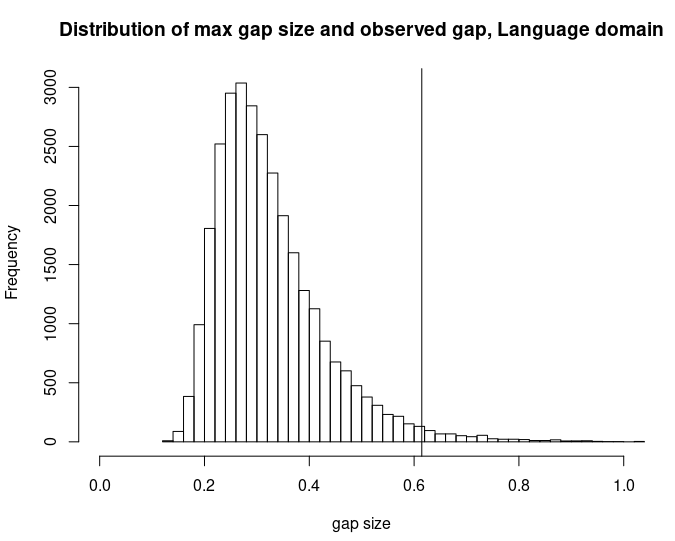}
      \caption{\small Distribution of maximum parameter gap observed in 10,000 simulations, fitting parameters only to models in the Other domain (left) or Language domain (right).}
      \label{fig:dmax_by_cat}
    \end{figure*}

\subsection{Further context}

We list now some facts that might be useful to better understand the gap.

\begin{enumerate}
    \item \textbf{The region in the gap is not completely empty}. We are aware of some models that lie within the gap, but which are not included in our datasets\footnote{This includes Anthropic’s 52B model, the 39B version of HyperCLOVA, the 68B version of FLAN, and the 62B version of PaLM.}. Some of these are smaller versions of larger models trained to test scaling laws, which we do not include because we only include the largest model from each paper. However, in our view, this does not explain why there are no publications whose biggest model falls into that range.

    We also found a single model, Anthropic’s 52B-parameter language model \cite{askell_anthropic}, which was not included in the dataset for not satisfying the inclusion criteria. Nevertheless, we think the relative scarcity of models in this range is still surprising, even if some models do lie within.
    
    \item \textbf{It is mostly a gap in language models}. Since only models in the Language and Other categories have reached the requisite size to be above the gap, we can restrict our analysis to these two domains. The same statistical analysis of the previous section, when performed only on systems in the Other domain, shows that the gap can be explained by random noise, whereas for systems in the Language domains the gap is very unlikely: The probability of observing this gap conditional on our model being correct is 0.23 for systems in the \textit{Other} domain and 0.02 for those in the \textit{Language} domain\footnote{There are 71 language models in total, of which 12 are above the gap.}. (see Figure \ref{fig:dmax_by_cat}). All of this suggests that we focus our analysis on language models.
    
    \item \textbf{The gap shrinks over time}\footnote{Concretely, it initially spanned a full order of magnitude, then was reduced to 0.91 OOMs by FLAN in 2021, then to 0.61 OOMs by Chinchilla, and finally to 0.54 OOMs by GPT-NeoX-20B in 2022.}. It seems likely that within the next couple of years it will disappear, as other domains catch up in model size and hardware and economic conditions change, and as models are trained for a wider range of purposes.
    
    \item \textbf{The first language model to surpass the gap was GPT-3}, which when released was 10 times bigger than the largest previous language models. There are around 4 jumps of this size for language models in our dataset\footnote{The other jumps were due to the release of Word2Vec, Hiero and IBM Watson. Of these, only Word2Vec and GPT-3 happened after 2010}.
    
    \item \textbf{The parameter gap probably reflects a divide between expensive models and cheaper models}. We should expect the parameter gap to reflect a proportional gap in training compute\footnote{Model size and training cost are not in general directly proportional, since the size of the training dataset also influences the amount of compute spent on training. However, increased understanding of scaling laws since 2020 has compelled researchers to use a nearly fixed ratio of model and dataset size.}. This is not true in general, but we do see a small gap if we focus only on language models (see Figure \ref{fig:compute}).
    
    Furthermore, if we plot training compute and model size, we observe two distinct clusters after GPT-3\footnote{Here we ignore Mixture of Experts (MoE) models because their scaling properties are very different from dense models.}, whose centers vary both across compute and across size (see Figure \ref{fig:compute})\footnote{We do not have estimates of the training compute for all the models, so this plot only includes a subset of the models.}. So we can probably say that the parameter gap reflects a divide between expensive models and cheaper models.
    
    \begin{figure*}[ht]
      \small
      \centering
      \includegraphics[width=0.48\textwidth]{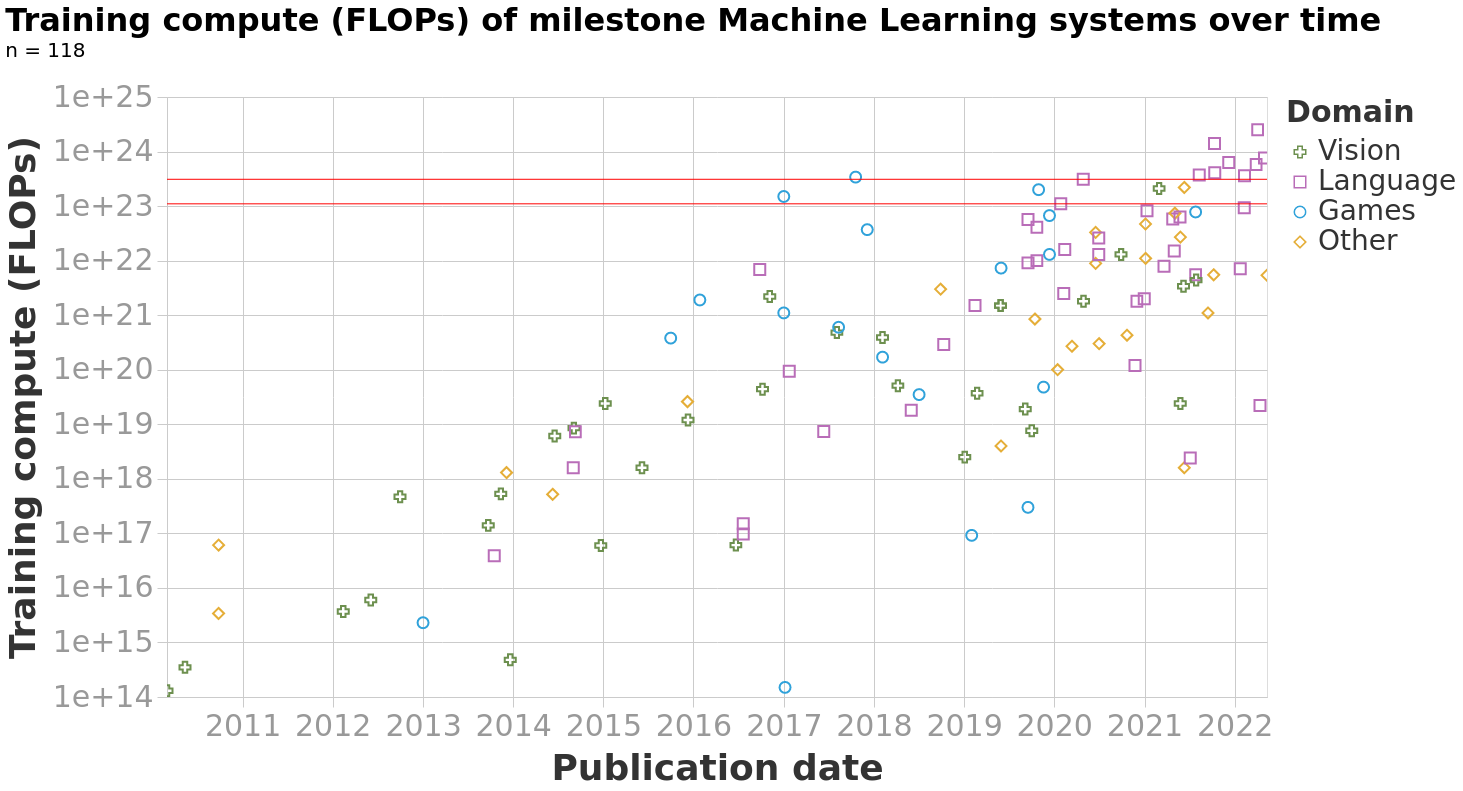}
      \includegraphics[width=0.42\textwidth]{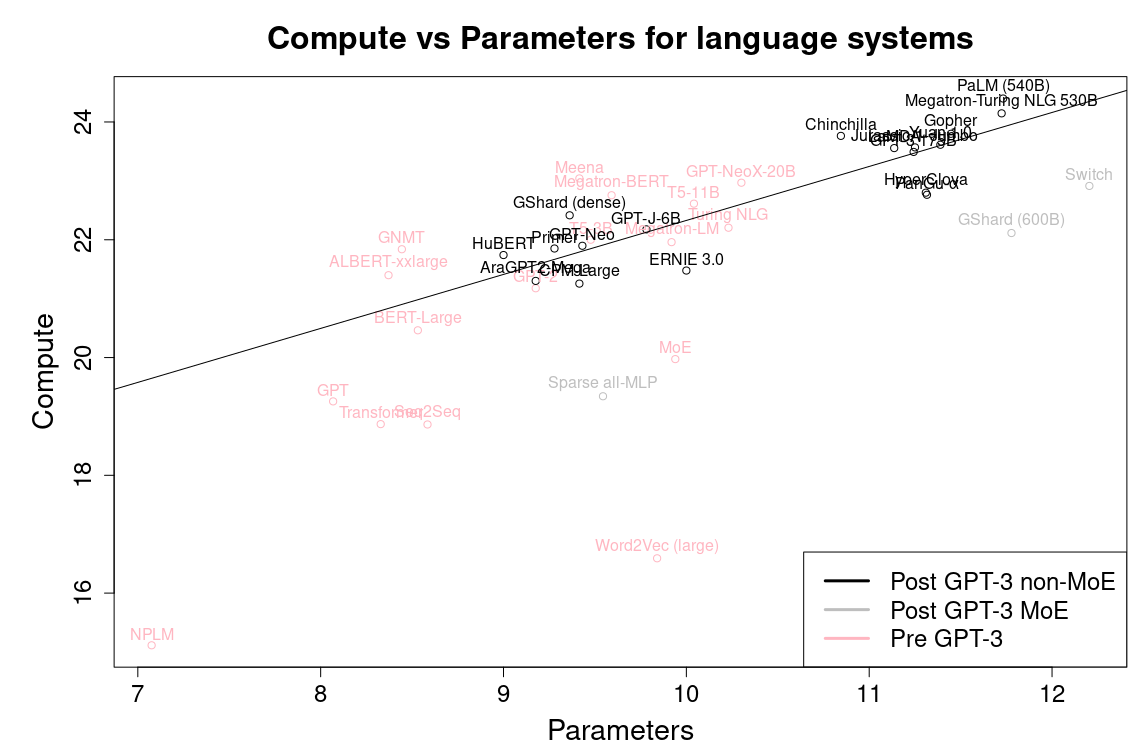}
      \caption{\small Left: Gap in training compute for language models, corresponding to the gap in parameters. Right: Training compute and model size of language models, showing two clusters of models after GPT-3. }
      \label{fig:compute}
    \end{figure*}
    
    \item \textbf{Models above the gap are almost exclusively language models, while those below it span all domains}. Models above the gap are overwhelmingly notable for pushing the state of the art in language generation either in English or Chinese, whereas models below the gap are much more varied in their domains and contributions (see Table~\ref{tab:table_hardware} in the appendix).
        
\end{enumerate}

\subsection{Possible explanations}
    
We present some hypotheses that try to explain the relative scarcity of models in the 20B-70B range. These theories can be classified into two groups:

\begin{enumerate}
    \item A “fundamental threshold” group of theories. These state that there is something specific about that size range that causes some tendency to avoid models within it. An example of this kind of theory is “training models at that scale is significantly less cost-effective than other scales”.
    
    \item An “external effects” group of theories that state that there is nothing particular about that range, and instead there are outside incentives like competitive pressure or different access to compute which produce this divide.
\end{enumerate}

In addition, there’s the possibility that there is not really a gap and instead our dataset is missing the relevant models. While this is somewhat plausible, we could not find many systems within that range outside our dataset.

\subsubsection{Fundamental threshold theories}

The fundamental threshold theories posit that there is some transition around the size of the gap where previous scaling strategies give way to new ones, in a way that training a model at the intermediate regime is not as cost-effective as training one on either side of the gap.

\textbf{Cloud vs bespoke infrastructure}: The first hypothesis from Derek Jones, was that ~50B parameters is the size at which it becomes more cost-effective to employ bespoke computing infrastructure (supercomputers)\footnote{Jones did not clarify exactly what he meant by bespoke infrastructure, but the way we understand it it means hardware which is in general not easily available for rental and that might have been configured specifically for the task at hand. An example might be BigScience’s language model being trained on the Jean Zay supercomputer, which is only available for academic research and has a team of engineers ready to help users.}, rather than generic cloud hardware.

It’s true that some of the bigger models have been trained on supercomputers, in particular BigScience’s language model as well as some of the Chinese models. However, we are not aware of other models that were trained in supercomputers and in general, this does not seem to be the case. Most of the systems above and below the gap were trained on TPUs and NVIDIA GPUs which are commercially available in cloud services. See Table~\ref{tab:table_hardware} in the appendix for a full list.

\textbf{Training parallelism threshold}: GPUs have a limited amount of memory, 32GB for NVIDIA V100, 40GB or 80GB for NVIDIA A100. Using a standard Adam optimizer with no model parallelism or optimizer state sharding, this amount of memory can fit around 1-2B parameters for training \cite{nvidia_large_scale}, since in addition to the parameters themselves the optimization process imposes additional memory overhead.

For this reason, models above that size usually use either pipeline parallelism or tensor parallelism to split the training onto multiple GPUs. Since a single computing node usually contains 8 or 16 GPUs\footnote{For example, NVIDIA’s DGX node contains 8 GPUs, whereas their DGX-2 node contains 16 GPUs.}, using this parallelism it is possible to train models up to around 20B parameters within a single node \cite{nvidia_large_scale,microsoft_zero_infty}.

To train bigger models, we know of two techniques: taking advantage of non-GPU memory to fit larger models in one node, and splitting the model across nodes. As far as we know, the first option incurs a significant decrease in the hardware utilization rate that can be attained. However, the second option requires significant code refactoring and engineering expertise, since this splitting requires significant communication between nodes, which are usually connected by a slower network than within-node GPUs. Moreover, once past this multi-node splitting threshold, training becomes more efficient with larger models \cite{nvidia_large_scale}.

So, projects passing this threshold must pay a fixed cost in engineering effort for training models bigger than 20B parameters. This implies that the marginal cost of going from 20B to, for example, 40B parameters might be higher than going from 40B to 60B parameters, so the total cost per parameter is much higher in the region immediately after the threshold.

We are quite confident from our exploration of the scaling literature that this threshold exists and it imposes a fixed cost, but it is not clear whether such a cost is high enough to explain the gap. In our correspondence with practitioners, this threshold has not been explicitly mentioned as a major factor in their scaling decisions, which suggests that it is not that important in practice.

    \textbf{Inference parallelism threshold}: Another possibility is that the parallelism threshold is more relevant during inference. Without needing to store optimizer data, a single GPU can fit a 20B size model, but not a 40B size one. In correspondence, EleutherAI stated this as a reason for limiting their model size to 20B.

Deployment usually consumes much more compute than training: Patterson et al. find that around 90\% of total ML compute is spent on inference \cite{patterson2021}, so it might make sense to optimize models for efficiency during inference more than training. However, from our understanding parallelizing inference poses less complex engineering challenges than training, since information only flows forward so pipeline parallelism can be used much more efficiently, reducing the need for other scaling techniques\footnote{We are not practitioners so we are not entirely sure about this, but basically during training the need to do a forward pass of activations and a backward pass of gradients leads to less utilization when using pipeline parallelism. However, if there is only a forward pass pipeline parallelism should be able to achieve high hardware utilization.}. So the cost of passing this threshold seems much lower than for training.

In our reading of the literature, we found quite a few papers that provide details on their training infrastructure setup, whereas the inference infrastructure setup is not usually mentioned. This provides some evidence that inference is not seen as important from an engineering standpoint.

For collectives such as EleutherAI whose goal is increasing access to cutting-edge language models, rather than commercial viability, reducing the number of GPUs needed during inference might be an important goal in itself. On the other hand, OPT-175B was also released to democratize access to big language models, yet it employs 16 V100 GPUs during deployment \cite{zhang_opt}. BigScience also has democratization as their goal, but they are training a 176B model which will also require parallelization during inference. This indicates that inference parallelization might not be such an important factor, even for actors motivated by democratization.

\subsubsection{External effects theories}

The external effects theories attempt to explain the gap by using arguments unrelated to the technical details of the 20B-70B parameter range, such as historical contingencies, economic or research incentives, and access to hardware.

\textbf{Compute divide + scaling laws}: Due to scaling laws, recent language models have been trained following a fixed ratio of training compute to model size \cite{Hoffmann2022}. So perhaps the reason for the gap lies in the ‘compute divide’ \cite{ahmed_compute_divide}: some players like Google or OpenAI have access to enormous computational resources, whereas academic laboratories or other researchers usually have fewer resources.

Under this hypothesis, large corporations and research laboratories could train bigger models and other researchers would be forced to work with smaller models. However, a significant number of models below the gap were produced by top research laboratories, and some models above the gap have been produced by other players\footnote{Examples of systems below the gap made by big players include Retro-7B, DeepNet and BigSSL. Systems above the gap are almost completely produced by corporations, but BigScience is training a 176B model which when completed will be above the gap. See Table \ref{tab:table_hardware} in the appendix for a list of models and the organization that created them.}. Thus, even if this divide exists and prevents less resourceful actors from training big models, it does not explain why no models trained by big corporations are present within the gap.

\textbf{Initial discontinuity + Record-setting models}: Since GPT-3 demonstrated the robustness of scaling, any project that wants to compete with it and push the state of the art in language generation must either be bigger or leverage new insights (for example, Chinchilla’s compute-optimal training). However, when GPT-3 was released it was one order of magnitude bigger than any previous language model. So its release caused a discontinuity in model size that was preserved by the incentives to outperform GPT-3.

Under this hypothesis, the models below the gap can not be directly competing with those above, perhaps because they are specialized in more niche tasks (like non-English languages), or because the researchers have few incentives to outcompete existing models (this might be the case for research collectives like BigScience or EleutherAI).

Indeed, as can be seen in Table \ref{tab:table_hardware}, most models above the gap are notable for having pushed the state of the art in English or Chinese language generation, whereas below the gap we find other tasks such as language understanding and Arabic and Japanese language generation. We also find more papers whose main contributions are algorithmic improvements or which were produced by independent authors or research collectives.

A problem with this hypothesis is that it predicts gaps that are not present: if this hypothesis was true we would expect to see a gap in our dataset every time there is a significant jump in model size. For example, when Word2vec was released it was an order of magnitude bigger than previous language models, but no gap appeared afterwards. Some peculiarities of the 2020 jump that might explain the presence of a gap are the increased importance of scaling laws and the fact that this time the jump represented a significant increase in training cost.

\subsubsection{Comparing hypotheses}

While we do not have evidence to establish whether our hypothesis fully explain the gap, we have outlined the reasons to reject some of them, as shown in the previous section. All in all, the hypotheses that we can’t rule out are the training and inference parallelism threshold and record-setting models.

More concretely, while the parallelism thresholds are well documented, we are uncertain about the fixed cost imposed by such thresholds and whether this cost is high enough to explain the gap. Especially for inference, where the cost of passing the threshold seems lower.

As for the record-setting models hypothesis, we have confirmed that most models below the gap released after GPT-3 do not seem to compete directly with big models on language generation performance, but instead make progress in niche tasks or introduce innovations that do not include raising the SOTA in language generation benchmarks (again, see Table \ref{tab:table_hardware}).

This makes the record-setting models hypothesis the one best supported by the evidence so far, but we still are not confident about its ability to fully account for the gap. Of course, these hypotheses are not mutually exclusive, and it is likely that all of them play some role in generating the gap. Establishing the relative importance of each of them is beyond our scope at this point.

\section{Conclusion}
\label{sec:conclusion}

In this article, we explored the model size of some notable machine learning systems and presented and analyzed some trends that this metric has been following over the years.

While this data is somewhat noisy and hard to reason about (for example, due to architectural differences between models), we believe we can draw from this data some conclusions about the dynamics of progress in ML from it:
\begin{itemize}
    \item While language and recommendation models have driven a huge increase in size over the last two years, other domains have seen much more moderate increases.
    \item If we restrict ourselves to the models above the gap, there are no signs of growth in model size in the past few years, so the overall trend will likely catch up in the next few years.
\end{itemize}

In addition, we can generate some predictions from our hypotheses on the parameter gap:
\begin{itemize}
    \item Hardware limitations might cause discontinuities or bottlenecks in scaling by limiting model size rather than compute. For example, 3,072 GPUs can be used to train a trillion parameter Transformer \cite{nvidia_large_scale}. Summit, the 4th top supercomputer as of July 2022, has 27,648 GPUs. So training dense language models beyond 10T parameters will likely require using non-GPU memory, bigger supercomputers (which take time to build), or communication between supercomputers.
    \item After a sudden jump in the largest model size used for a given task or benchmark, we should expect to stop seeing smaller notable models competing in  the same tasks or benchmarks, at least in domains with robust scaling laws. This implies that a single well-funded actor can suddenly change the dynamics of AI research. In particular, by (a) causing less resourceful actors to stop working on a given task or benchmark since the well-funded actor jumped to too large a model size to be worth trying to compete with, (b) causing other well funded actors to increase their spending if they want to remain competitive.
\end{itemize}

\section*{Acknowledgments}
\label{sec:acknowledgments}

Many thanks to Derek Jones who spotted the parameter gap, as well as Stella Biderman, Peter Wildeford, Yu Zhang, Furu Wei, Stephen Roller, Laurent Sifre and Jared Kaplan for answering our questions. We also thank Ben Cottier, Michael Ashcroft and Michael Aird for their reviews and comments on earlier drafts of this paper.

\nocite{*}
\bibliography{IEEEabrv,references}

\clearpage
\onecolumn

\section*{Appendix: list of models}
\label{sec:appendix}

\begin{table*}[ht]
\centering
\begin{tabular}{|lllll|}
\hline
\multicolumn{5}{|c|}{\textbf{Above gap}}                                                                                                                                                                        \\ \hline
\multicolumn{1}{|l|}{\textbf{System}}                                & \multicolumn{1}{l|}{\textbf{Organization}} & \multicolumn{1}{l|}{\textbf{Hardware type}} & \multicolumn{1}{l|}{\textbf{Number of chips}}         & \multicolumn{1}{l|}{\textbf{Notability}}                        \\ \hline
\multicolumn{1}{|l|}{GPT-3}                                          & \multicolumn{1}{l|}{OpenAI}                & \multicolumn{1}{l|}{V100}   
& \multicolumn{1}{l|}{} & Language generation SOTA                   \\ \hline
\multicolumn{1}{|l|}{FLAN}                                           & \multicolumn{1}{l|}{Google}                & \multicolumn{1}{l|}{TPUv3} & \multicolumn{1}{l|}{128}                 & Language generation SOTA                   \\ \hline
\multicolumn{1}{|l|}{ERNIE 3.0 Titan}                                & \multicolumn{1}{l|}{Baidu}                 & \multicolumn{1}{l|}{V100 + Ascend 910} & \multicolumn{1}{l|}{} & Language generation SOTA (Chinese)         \\ \hline
\multicolumn{1}{|l|}{Efficient Large-Scale Training on GPU Clusters} & \multicolumn{1}{l|}{Nvidia}                & \multicolumn{1}{l|}{A100}        &  \multicolumn{1}{l|}{3072} &  Scaling                                    \\ \hline
\multicolumn{1}{|l|}{Jurassic-1}                                     & \multicolumn{1}{l|}{AI21}                  & \multicolumn{1}{l|}{GPU}                  & \multicolumn{1}{l|}{800} & Language generation SOTA                   \\ \hline
\multicolumn{1}{|l|}{Yuan 1.0}                                       & \multicolumn{1}{l|}{Inspur}                & \multicolumn{1}{l|}{GPU (likely V100)}   &  \multicolumn{1}{l|}{2128} & Language generation SOTA (Chinese)         \\ \hline
\multicolumn{1}{|l|}{Turing-NLG 530B}                                & \multicolumn{1}{l|}{Microsoft}             & \multicolumn{1}{l|}{A100} & \multicolumn{1}{l|}{$\sim$8000} & Language generation SOTA                   \\ \hline
\multicolumn{1}{|l|}{Gopher}                                         & \multicolumn{1}{l|}{Google}                & \multicolumn{1}{l|}{TPUv3}                & \multicolumn{1}{l|}{4096} & Language generation SOTA                   \\ \hline
\multicolumn{1}{|l|}{LaMDA}                                          & \multicolumn{1}{l|}{Google}                & \multicolumn{1}{l|}{TPUv3}                & \multicolumn{1}{l|}{1024} & Dialog                                     \\ \hline
\multicolumn{1}{|l|}{PaLM}                                           & \multicolumn{1}{l|}{Google}                & \multicolumn{1}{l|}{TPUv4}                & \multicolumn{1}{l|}{6144} & Language generation SOTA                   \\ \hline
\multicolumn{1}{|l|}{Flamingo}                                       & \multicolumn{1}{l|}{Google}                & \multicolumn{1}{l|}{TPUv4}                & \multicolumn{1}{l|}{1536} & Multimodal text + image understanding SOTA \\ \hline
\multicolumn{1}{|l|}{OPT-175B}                                       & \multicolumn{1}{l|}{Meta}                  & \multicolumn{1}{l|}{A100}                  & \multicolumn{1}{l|}{992} & Efficiency, democratization                \\ \hline
\multicolumn{1}{|l|}{Pangu-$\alpha$}                                        & \multicolumn{1}{l|}{Pangu-$\alpha$ team}          & \multicolumn{1}{l|}{Ascend 910}           & \multicolumn{1}{l|}{2048} & Language generation SOTA (Chinese)         \\ \hline
\multicolumn{1}{|l|}{HyperCLOVA}                                     & \multicolumn{1}{l|}{Naver}                 & \multicolumn{1}{l|}{A100}        & \multicolumn{1}{l|}{1024} & Language generation SOTA (Korean)          \\ \hline
\end{tabular}
\end{table*}

\begin{table*}[ht]
  \centering
\begin{tabular}{|lllll|}
\hline
\multicolumn{5}{|c|}{\textbf{Below gap}}                                                                                                                                                 \\ \hline
\multicolumn{1}{|l|}{\textbf{System}}              & \multicolumn{1}{l|}{\textbf{Org}}                  & \multicolumn{1}{l|}{\textbf{Hardware type}} & 
\multicolumn{1}{l|}{\textbf{Number of chips}} &
\textbf{Notability}                   \\ \hline
\multicolumn{1}{|l|}{DeepNet}                      & \multicolumn{1}{l|}{Microsoft}                     & \multicolumn{1}{l|}{}     & \multicolumn{1}{l|}{}        & Algorithmic innovation                \\ \hline
\multicolumn{1}{|l|}{Retro-7B}                     & \multicolumn{1}{l|}{DeepMind}                      & \multicolumn{1}{l|}{TPUv3}  &  \multicolumn{1}{l|}{}           & Algorithmic innovation                \\ \hline
\multicolumn{1}{|l|}{GPT-J-6B}                     & \multicolumn{1}{l|}{EleutherAI}                    & \multicolumn{1}{l|}{TPUv3}             &   \multicolumn{1}{l|}{} &  Produced by research collective       \\ \hline
\multicolumn{1}{|l|}{Seq2seq}                      & \multicolumn{1}{l|}{Facebook}                      & \multicolumn{1}{l|}{GPUs}              &  \multicolumn{1}{l|}{} &  Dialog SOTA                           \\ \hline
\multicolumn{1}{|l|}{BigSSL}                       & \multicolumn{1}{l|}{Google}                        & \multicolumn{1}{l|}{TPUv3}        &  \multicolumn{1}{l|}{1024} &  Speech recognition SOTA               \\ \hline
\multicolumn{1}{|l|}{Turing NLG 17B\footnote{This model was released before GPT-3.}}               & \multicolumn{1}{l|}{Microsoft}                     & \multicolumn{1}{l|}{V100}  &  \multicolumn{1}{l|}{4096} &  Language generation SOTA              \\ \hline
\multicolumn{1}{|l|}{DeBERTa}                      & \multicolumn{1}{l|}{Microsoft}                     & \multicolumn{1}{l|}{V100}  &  \multicolumn{1}{l|}{4096} &  Language understanding SOTA           \\ \hline
\multicolumn{1}{|l|}{Japanese dialog transformers} & \multicolumn{1}{l|}{NTT}                           & \multicolumn{1}{l|}{V100 16GB}     &  \multicolumn{1}{l|}{400} &  Dialog (Japanese)                     \\ \hline
\multicolumn{1}{|l|}{Primer}                       & \multicolumn{1}{l|}{Google}                        & \multicolumn{1}{l|}{TPUv4\footnote{This is the hardware used for the biggest model, other models used TPUv2, TPUv3 and V100 GPUs}}         &  \multicolumn{1}{l|}{512} &  Algorithmic innovation                \\ \hline
\multicolumn{1}{|l|}{AraGPT2}                      & \multicolumn{1}{l|}{American University of Beirut} & \multicolumn{1}{l|}{TPUv3}         &  \multicolumn{1}{l|}{128} &   Language generation (Arabic)          \\ \hline
\multicolumn{1}{|l|}{T0-XXL}                       & \multicolumn{1}{l|}{BigScience}                    & \multicolumn{1}{l|}{TPUv3}         &  \multicolumn{1}{l|}{512} &   Improved training methodology         \\ \hline
\multicolumn{1}{|l|}{XLM-R}                        & \multicolumn{1}{l|}{Facebook}                      & \multicolumn{1}{l|}{}       &     \multicolumn{1}{l|}{}        & Language understanding SOTA           \\ \hline
\multicolumn{1}{|l|}{Codex}                        & \multicolumn{1}{l|}{OpenAI}                        & \multicolumn{1}{l|}{}       &     \multicolumn{1}{l|}{}           & Code generation SOTA                  \\ \hline
\multicolumn{1}{|l|}{ERNIE 3.0}                    & \multicolumn{1}{l|}{Baidu}                         & \multicolumn{1}{l|}{V100}    &     \multicolumn{1}{l|}{384}      & Language understanding SOTA (Chinese) \\ \hline
\multicolumn{1}{|l|}{GPT-NeoX-20B}                 & \multicolumn{1}{l|}{EleutherAI}                    & \multicolumn{1}{l|}{A100 40GB}   &     \multicolumn{1}{l|}{96}   & Produced by research collective       \\ \hline
\end{tabular}
  \vspace{.5em}
  \caption{List of post-GPT-3 language models above and below the gap, together with the organization which produced them, the hardware used for training, and the reason why they are notable according to our criteria. This last item was quickly and subjectively evaluated from quick glances at the paper abstracts, so it’s not a complete characterization of the merits of each paper.}
  \label{tab:table_hardware}
\end{table*}

\end{document}